\documentclass[preprint, prX]{revtex4}

\usepackage{amsmath}    
\usepackage{graphicx}   
\usepackage{verbatim}   
\usepackage{color}      
\usepackage{subfigure}  
\usepackage{hyperref}   
\usepackage{amssymb}    
\usepackage{epsfig}
\usepackage{graphics,graphicx}
\usepackage{setspace}
\usepackage{url}
\usepackage{algorithm,algorithmic}

\begin{document}

\begin{abstract}
This paper introduces a novel idea for representation of individuals using quaternions in swarm intelligence and evolutionary algorithms. Quaternions are a number system, which extends complex numbers. They are successfully applied to problems of theoretical physics and to those areas needing fast rotation calculations. We propose the application of quaternions in optimization, more precisely, we have been using quaternions for representation of individuals in Bat algorithm. The preliminary results of our experiments when optimizing a test-suite consisting of ten standard functions showed that this new algorithm significantly improved the results of the original Bat algorithm. Moreover, the obtained results are comparable with other swarm intelligence and evolutionary algorithms, like the artificial bees colony, and differential evolution. We believe that this representation could also be successfully applied to other swarm intelligence and evolutionary algorithms.

\textit{To cite paper as follows: I. Fister, I. Fister Jr.: Using the quaternion's representation of individuals in swarm intelligence and evolutionary computation, {\em Technical Report on Faculty of Electrical Engineering and Computer Science}, Maribor, Slovenia, 2013.
}

\end{abstract}

\title{Using the quaternion's representation of individuals in swarm intelligence and evolutionary computation}

\author{Iztok Fister}
\altaffiliation{University of Maribor, Faculty of Electrical Engineering and Computer Science
Smetanova 17, 2000 Maribor}
\email{iztok.fister@uni-mb.si}

\author{Iztok Fister Jr.}
\altaffiliation{University of Maribor, Faculty of Electrical Engineering and Computer Science
Smetanova 17, 2000 Maribor}
\email{iztok.fister@guest.arnes.si}

\maketitle

\section{Introduction}
Human desire has always been to build an automatic problem solver that would be able to solve any kind of problems within mathematics, computer science and engineering. In order to fulfil this desire, humans have often resorted to solutions from Nature that are eternal sources of inspiration. Let us mention only two the more powerful nature-inspired problem solvers nowadays:
\begin{itemize}
\item \textit{Evolutionary Algorithms} (EA) inspired by the Darwinian evolutionary process, where in nature the fittest individuals have the greater possibilities for survival and pass-on their characteristics to their offspring during a process of reproduction~\cite{Darwin:1859}.
\item \textit{Swarm Intelligence} (SI) inspired by the collective behaviour within self-organized and decentralized natural systems, e.g., ant colony, bees, flocks of birds or shoals of fish~\cite{Blum:2008}.
\end{itemize} 

Evolutionary computation (EC) is a contemporary term that denotes the whole field of computing inspired by Darwinian principles of natural evolution. As a result, evolutionary algorithms which are involved in this field are divided into the following disciplines: genetic algorithms (GA)~\cite{Goldberg:1989,Holland:1992}, evolution strategies (ES)~\cite{baeck:1996}, differential evolution (DE)~\cite{storn1997differential,brest2006self,das2011differential}, evolutionary programming (EP)~\cite{fogel:1966}, and genetic programming (GP)~\cite{Koza:1994}. Although all these disciplines were developed independently, they share similar characteristics when solving problems. Evolutionary algorithms have been applied in wide-areas of optimization, modeling, and simulation. 

Swarm intelligence is an artificial intelligence (AI) discipline concerned with the design of intelligent multi-agent systems that was inspired by the collective behaviour of social insects like ants, termites, bees, and wasps, as well as from other animal societies like flocks of bird or shoals of fish~\cite{Blum:2008}. Algorithms from this field have been applicable primarily for optimization problems, and the control of robots. The more notable swarm intelligence disciplines are as follows: Ant Colony Optimization (ACO)~~\cite{Dorigo:1999,Korosec:2012}, Particle Swarm Optimization (PSO)~\cite{kennedy1995particle}, Artificial Bees Colony optimization (ABC)~\cite{Karaboga:2007,Fister:2012}, Firefly Algorithm (FA)~\cite{yang2008afirefly}, Cuckoo Search (CS)~\cite{Yang:2009}, Bat Algorithm (BA)~\cite{Yang:2010}, etc.

This paper focuses mainly on optimization problems. Typically, optimization algorithms do not search for solutions within the \textit{original problem context} because these represent the problems on computers by means of data structures. For instance, genetic algorithms use binary representation of individuals, evolution strategy and differential evolution real numbers, genetic programming programs in Lisp, and evolutionary programming finite state automata. In general, algorithms in swarm intelligence represent solutions to problem as binary or real-valued vectors. 

When solving the optimization problem using genetic algorithms, some mapping between \textit{problem context} and binary represented \textit{problem-solving space} needs to be performed. That is that in the original problem context, the candidate solution determines so-named \textit{phenotype space}, i.e., set of points that form the space of possible solutions. On the other hand, the solution in the problem-solving space determines the set of points that forms the so-named \textit{genotype space}. By means of mapping from the genotype to the phenotype, the candidate solution is \textit{encoded} as a binary vector. In contrast, when the mapping from the phenotype to the genotype is taken into consideration, the candidate solution is \textit{decoded} from the binary representation of individuals. Swarm intelligence and evolutionary algorithms hold the representation of possible solutions within a \textit{population}. The population is a set of multiple copies of candidate solutions. 

The variation operators are applied in order to create new individuals from old ones. From the search perspective, the swarm intelligence and the evolutionary algorithms act according the 'generate-and-test' principle, where the variation operators perform the generate phase. However, variation operators operate differently in swarm intelligence and evolutionary algorithms. In general, algorithms in swarm intelligence support a \textit{move} variation operator that obtains a new position for a individual from the old one in a specific manner. For example, in particle swarm optimization (PSO)~\cite{kennedy1995particle} a new position for an individual depends on the global best and local best position of individuals within the swarm (i.e., in population). Typically, evolutionary algorithms support two variation operators that mimic operations in nature, i.e., \textit{mutation} and \textit{crossover}. Evolutionary algorithms simulate the process of natural selection using \textit{selection} operator. Actually, two selection operators exist. The former selects the parents for the reproduction (also \textit{parent selection}), whilst the latter determines the surviving offspring (also \textit{survivor selection}).  

The representation plays an important role in the performance and solution quality of the swarm intelligence or evolutionary methods. Therefore, the tasks of designing such programs are to find a \textit{proper representation} for the problem, and to develop \textit{appropriate search operators}~\cite{Rothlauf:2006}. The proper representation needs to encode all possible solutions of the optimization problem. On the other hand, the appropriate search operator should be applicable to the proper representation. In general, no theoretical methods exist nowadays for describing the effects of representation on the performance. The proper representation of a specific problem mainly depends on the intuition of the swarm intelligence or evolutionary designer. Therefore, developing a new representation is often a result of the repeated 'trial-and-error' principle.

This paper proposes a new representation for individuals using quaternions. In mathematics, the quaternions extend complex numbers. Quaternion algebra is connected with special features of the geometry of the appropriate Euclidean spaces. The idea of quaternions occurred to William Rowan Hamilton in 1845~\cite{Hamilton:1899} whilst he was walking along the Royal Canal to a meeting of the Royal Irish Academy in Dublin. Interestingly, he carved the fundamental formula of quaternion algebra, that is: 
\begin{equation}
i^{2}=j^{2}=k^{2}=ijk=-1,
\end{equation}

\noindent into the stone of the Brougham Bridge.

Quaternions are especially appropriate within those areas where it is necessary to compose rotations with minimal computation, e.g., programming the video games or controllers of spacecraft~\cite{Conway:2003}. For instance, 3-dimensional rotation can be specified by a single quaternion, whilst a pair of quaternions are needed for 4-dimensional rotation. The quaternion calculus is introduced in several physical applications, like: crystallography, the kinematics of rigid body motion, the Thomas precession, the special theory of relativity, and classical electromagnetism~\cite{Girard:1984}. A step forward in the popularization of quaternions was achieved by Joachim Lambek in 1995~\cite{Lambek:1995}, who stated that quaternions can provide a shortcut for pure mathematicians who wish to familiarize themselves with certain aspects of theoretical physics. 

As far as we know, quaternions have never been used in optimization. Each optimization search process depends on balancing between two major components: exploration and exploitation~\cite{Crepinsek:2011}. Both terms were defined implicitly and are affected by the algorithms' control parameters. For optimization  algorithms, the exploration denotes the process of discovering diverse solutions within the search space, whilst exploitation means focusing the search process within the vicinities of the best solutions, thus, exploiting the information discovered so far. As a result, too much exploration can lead to inefficient search, whilst too much exploitation can cause the premature convergence of a search algorithm where the search process, usually due to reducing the population diversity, can be trapped into a local optimum~\cite{Eiben:2003}. In place of premature convergence, a phenomenon of stagnation is typical for swarm intelligence, which can occur when the search algorithm cannot improve the best performance (also fitness) although the diversity is still high~\cite{Neri:2012}. 

In order to avoid the stagnation in swarm intelligence as well as the premature convergence in evolutionary algorithms, we propose the representation of individuals using quaternions. Rather than using the real-valued elements of individuals, the elements of the solution are represented by quaternions. In this manner, each an 1-dimensional element of solution is placed into a 4-dimensional quaternion space, where the search process acts using the operators of quaternion algebra. However, the quality of solution is evaluated in phenotype space, therefore, each 4-dimensional quaternion is mapped back into a 1-dimensional element using the quaternion's normalization function. During this mapping some information can be lost, but on the other hand, it is expected that the fitness landscape as described by quaternions will replace the flat areas and plateaus of the original fitness landscape with peaks and valleys that are more appropriated for exploration. 

Our intention in the future is to apply the quaternion's representation to some swarm intelligence algorithms, like BA, FA and PSO, and some evolutionary algorithms, like DE and ES, in order to show that this representation could have beneficial effects on the swarm intelligence and evolutionary search processes, especially by avoiding the stagnation and premature convergence.

\bigskip{\small \smallskip\noindent Updated 10 June 2013.}

\begin {thebibliography} {99}

\bibitem{Darwin:1859} Darwin, C., \textit{The origin of species.} John Murray, London, UK (1859).
\bibitem{Blum:2008} Blum, C. and Li, X., \textit{Swarm Intelligence in Optimization.} In: C. Blum and D. Merkle (eds.) Swarm Intelligence: Introduction and Applications, pp. 43-86. Springer Verlag, Berlin (2008).
\bibitem{Goldberg:1989} Goldberg, D., \textit{Genetic Algorithms in Search, Optimization, and Machine Learning.} Addison-Wesley, Massachusetts, US (1989).
\bibitem{Holland:1992} Holland, J.H., \textit{Adaptation in Natural and Artificial Systems: An Introductory Analysis with Applications to Biology, Control and Artificial Intelligence.} MIT Press, Cambridge, MA, US (1992).
\bibitem{baeck:1996} B{\"a}ck, T., \textit{Evolutionary algorithms in theory and practice - evolution strategies, evolutionary programming, genetic algorithms.} Oxford University Press, Oxford, UK (1996).
\bibitem{storn1997differential} Storn, R. and Price, K., \textit{Differential evolution--a simple and efficient heuristic for global optimization over continuous spaces.} Journal of global optimization, 4(11):341-359 (1997).
\bibitem{brest2006self} Brest, J. and Greiner, S. and Bo\v{s}kovi\'{c}, B. and Mernik, M. and \v{Z}umer, V., \textit{Self-adapting control parameters in differential evolution: A comparative study on numerical benchmark problems.} IEEE Transactions on Evolutionary Computation, 10(6):646-657 (2006).
\bibitem{das2011differential} Das, S. and Suganthan, P.N., \textit{Differential evolution: A survey of the state-of-the-art.} IEEE Transactions on Evolutionary Computation, 15(1):4-31 (2011).
\bibitem{fogel:1966} Fogel, L. J. and Owens, A. J. and Walsh, M. J., \textit{Artificial Intelligence through Simulated Evolution.} John Wiley, New York, US (1966).
\bibitem{Koza:1994} Koza, J.R., \textit{Genetic programming 2 - automatic discovery of reusable programs.} MIT Press,  Massachusetts, US (1994).
\bibitem{Dorigo:1999} Dorigo, M. and {Di Caro}, G, \textit{The Ant Colony Optimization meta-heuristic.} In: D. Corne and M. Dorigo and F. Glover (eds.) New Ideas in Optimization, pp. 11-32. McGraw Hill, London, UK (1999).
\bibitem{Korosec:2012} Peter Koro\v{s}ec and Jurij \v{S}ilc and Bogdan Filipi\v{c}, \textit{The differential ant-stigmergy algorithm.} Information Sciences, 192(0):82-97 (2012).
\bibitem{kennedy1995particle} Kennedy, J. and Eberhart, R., \textit{Particle swarm optimization.} In: Proceedings of IEEE International Conference on Neural Networks, pp. 1942-1948. (1995).
\bibitem{Karaboga:2007} Karaboga, D. and Basturk, B., \textit{A powerful and efficient algorithm for numerical function optimization: artificial bee colony {(ABC)} algorithm.} Journal of Global Optimization, 39(3):459-471 (2007).
\bibitem{Fister:2012} Fister, I. and Fister, I.Jr. and Brest, J. and \v{Z}umer, V., \textit{Memetic artificial bee colony algorithm for large-scale global optimization.} In: Proceedings of IEEE Congress on Evolutionary Computation, pp. 1-8. (2012).
\bibitem{yang2008afirefly} Yang, X.S., \textit{Firefly algorithm.} Nature-Inspired Metaheuristic Algorithms, Wiley Online Library, pp. 79-90 (2008).
\bibitem{Yang:2009} Yang, X.S. and Deb, S., \textit{Cuckoo search via Levy flights.} In: World Congress on Nature \& Biologically Inspired Computing (NaBIC 2009), pp. 210-214. (2009).
\bibitem{Yang:2010} Yang, X.S., \textit{A New Metaheuristic Bat-Inspired Algorithm.} In: C. Cruz and J.R. González and N. Krasnogor, D.A. Pelta, G. Terrazas (eds.) Nature Inspired Cooperative Strategies for Optimization (NISCO 2010), pp. 65-74. Springer Verlag, Berlin (2010).
\bibitem{Rothlauf:2006} Rothlauf, F., \textit{Representations for genetic and evolutionary algorithms.} Springer Verlag, Berlin (2006).
\bibitem{Lambek:1995} Lambek, J., \textit{If {H}amilton had prevailed: quaternions in physics.} The {M}athematical {I}ntelligencer, 17(4):7-15 (1995).
\bibitem{Girard:1984} Girard, P.R., \textit{The quaternion group and modern physics.} European {J}ournal of {P}hysics, European Physical Society, Northern Ireland 5:25-32 (1984).
\bibitem{Hamilton:1899} Hamilton, W.R., \textit{Elements of quaternions.} Longmans, Green and Co. (1899).
\bibitem{Conway:2003} Conway, J.H. and Smith, D.A., \textit{On Quaternions and Octonions: Their Geometry, Arithmetic, and Symmetry.} A K Reters, Wellesley, Massachusetts (2003).
\bibitem{Crepinsek:2011} \v{C}repin\v{s}ek, M. and Mernik, M. and Liu, S., \textit{Analysis of exploration and exploitations in evolutionary algorithms by ancestry trees.} International Journal of Innovative Computing and Applications, Inderscience 3(1):11-19 (2011).
\bibitem{Neri:2012} Neri, F., \textit{Diversity Management in Memetic Algorithms.} In: F. Neri and C. Cotta and P. Moscato (eds.) Handbook of Memetic Algorithms, 379:153-165. Springer Verlag, Berlin (2012).
\bibitem{Eiben:2003} Eiben, A.E. and Smith, J.E., \textit{Introduction to Evolutionary Computing.} Springer-Verlag,  Berlin (2003).

\end {thebibliography}
\end{document}